%% file: root.tex
\documentclass[letterpaper, 10 pt, conference]{ieeeconf}  

\IEEEoverridecommandlockouts                              
                                                          
\usepackage{amsmath} 
\usepackage{amsfonts}
\usepackage{graphicx}
\usepackage{subcaption}
\usepackage[table, dvipsnames]{xcolor} 
\usepackage{multirow}
\usepackage{mathtools}
\let\labelindent\relax

\usepackage{enumitem} 
\usepackage{booktabs}
\graphicspath{{figures/}}

\usepackage{tabularx}
\usepackage{threeparttable}
\usepackage{makecell}

\usepackage[pagewise]{lineno}

\usepackage{booktabs}
\usepackage[noadjust]{cite}
\nocite{*}  
\makeatletter
\let\NAT@parse\undefined
\makeatother

\usepackage{hyperref}
\hypersetup{
    colorlinks=true,
    citecolor=Green,
    linkcolor=NavyBlue,
    filecolor=OrangeRed,      
    urlcolor=OrangeRed,
    pdfpagemode=FullScreen,
}
\input{common}

\title{\LARGE \bf
    Learned Tree Search for Long-Horizon Social Robot Navigation in Shared Airspace
}

\author{Ingrid Navarro$^{1 \ast}$, Jay Patrikar$^{1 \ast}$, Joao P. A. Dantas$^{2}$, Rohan Baijal$^{3}$, Ian Higgins$^{1}$, \\ Sebastian Scherer$^{1}$ and Jean Oh$^{1}$
\thanks{$^{1}$Authors are with the Robotics Institute, Carnegie Mellon University, Pittsburgh, PA, USA. {\tt\scriptsize \{ingridn, jaypat, ihiggins, basti, jeanoh\}@cs.cmu.edu}}
\thanks{$^{2}$Author is with the Institute for Advanced Studies, Sao Jose dos Campos, SP, Brazil. Work done as a visiting researcher at Carnegie Mellon University. {\tt\scriptsize dantasjpad@fab.mil.br}}
\thanks{$^{3}$Author is with the Indian Institute of Technology Kanpur, Kanpur, Uttar Pradesh, India. Work done as a visiting researcher at Carnegie Mellon University. {\tt\scriptsize rbaijal@iitk.ac.in}}
\thanks{$^{\ast}$Equal contribution.}
}

\begin{document}

\maketitle
\thispagestyle{empty}
\pagestyle{empty}

\begin{abstract}
The fast-growing demand for fully autonomous aerial operations in shared spaces
necessitates developing trustworthy agents that can safely and seamlessly navigate in crowded, dynamic spaces. 
In this work, we propose Social Robot Tree Search (\sorts), an algorithm for the safe navigation of mobile robots in social domains. \sorts~aims to augment existing socially-aware trajectory prediction policies with a Monte Carlo Tree Search planner for improved downstream navigation of mobile robots. 
To evaluate the performance of our method, we choose the use case of social navigation for general aviation. To aid this evaluation, within this work, we also introduce \xplaneros, a high-fidelity aerial simulator, to enable more research in full-scale aerial autonomy. 
By conducting a user study based on the assessments of 26 FAA-certified pilots, we show that \sorts~performs comparably to a competent human pilot, significantly outperforming our baseline algorithm. We further complement these results with self-play experiments, showcasing our algorithm's behavior in scenarios with increasing complexity. 
\href{https://github.com/cmubig/sorts}{[Code]}\footnote{ \href{https://github.com/cmubig/sorts}{Codebase: https://github.com/cmubig/sorts}}$\mid$  
\href{https://youtu.be/PBE3O4cW2rI}{[Video]}\footnote{ \href{https://youtu.be/PBE3O4cW2rI}{Video: https://youtu.be/PBE3O4cW2rI}}  
\end{abstract}

\input{sections/introduction}

\input{sections/related_works}

\input{sections/approach}

\input{sections/experiment_setup} 

\input{sections/results}

\input{sections/conclusion}


\section*{ACKNOWLEDGMENT}

This work was supported by the Army Futures Command Artificial Intelligence Integration Center (AI2C), the Ministry of Trade, Industry and Energy (MOTIE), the Korea Institute of Advancement of Technology (KIAT), and the Brazilian Air Force. The views expressed in this article do not necessarily represent those of the aforementioned entities. We thank Jasmine Jerry Aloor for her support. We also thank the pilots from Condor Aero Club (KPJC) and ABC Flying Club (KAGC) for their participation in our user study. 



\bibliography{ref}

\end{document}

%% file: common.tex
\usepackage{graphicx}
\usepackage{hyperref}
\usepackage{keyval}
\usepackage{algorithm}
\usepackage{algpseudocode}
\usepackage{amsmath}
\usepackage{amssymb}
\usepackage{cleveref}
\usepackage{fancyhdr} 
\usepackage{color}
\usepackage{transparent}
\newcommand{\idest}{{\it i.e.}, }
\newcommand{\exempli}{{\it e.g.}, }

\newcommand{\sorts}{{SoRTS}}
\newcommand{\trajair}{{TrajAir}}
\newcommand{\xplaneros}{{X-PlaneROS}}
\newcommand{\sprnn}{{Social-PatteRNN}}
\newcommand{\rosplane}{{ROS-Plane}}

\newcommand{\state}{{\mathbf s}}

\newcommand{\action}{{\mathbf a}}
\newcommand{\R}{{\mathbb R}}

\newcommand{\modref}{{\textcolor{Orchid!80}{Reference Module}}}
\newcommand{\modval}{{\textcolor{Blue!90}{Cost Map}}}
\newcommand{\modsoc}{{\textcolor{Red!90}{Social Module}}}

\newcommand{\alghuman}{{\textcolor{Melon}{Human}}}
\newcommand{\algablation}{{\textcolor{Green}{Baseline}}}
\newcommand{\algsorts}{{\textcolor{RoyalBlue}{SoRTS}}}
\newcommand{\user}{{\textcolor{Magenta}{User}}}

\usepackage{booktabs}

\usepackage{xargs} 
\usepackage[colorinlistoftodos,prependcaption,textsize=tiny]{todonotes}
\newcommandx{\unsure}[2][1=]{\todo[linecolor=red,backgroundcolor=red!25,bordercolor=red,#1]{#2}}
\newcommandx{\change}[2][1=]{\todo[linecolor=blue,backgroundcolor=blue!25,bordercolor=blue,#1]{#2}}
\newcommandx{\info}[2][1=]{\todo[linecolor=OliveGreen,backgroundcolor=OliveGreen!25,bordercolor=OliveGreen,#1]{#2}}
\newcommandx{\improvement}[2][1=]{\todo[linecolor=Plum,backgroundcolor=Plum!25,bordercolor=Plum,#1]{#2}}
\newcommandx{\thiswillnotshow}[2][1=]{\todo[disable,#1]{#2}}
 
\def\HiLiYellow{\leavevmode\rlap{\hbox to \hsize{\color{yellow!20}\leaders\hrule height .8\baselineskip depth .4ex\hfill}}}
  
\fancypagestyle{firststyle}
{
  \fancyhf{}

}

\fancypagestyle{secondstyle}
{
  \fancyhf{}
  \fancyhead[R]{\footnotesize \thepage}

}

\usepackage{cleveref}

%% file: sections/introduction.tex
\section{Introduction} \label{sec:introduction}

A social robot strives to synthesize decision policies that enable it to seamlessly interact with humans, ensuring social compliance while attaining its desired goal. While marked progress has been made in social navigation and motion prediction \cite{mavrogiannis2021core, rudenko2020human, tian2022safety}, achieving seamless navigation among humans while balancing social and self-interested objectives remains challenging.
Social navigation can be formulated as a Partially Observable Stochastic Game (POSG) \cite{sadigh2018planning, riviere2021neural}. Deep Reinforcement Learning (DRL) methods \cite{matsuzaki2022learning} explicitly formulate the POSG reward to derive a policy from simulated self-play experiments. While such techniques are promising in sparse-data domains where the robot is easily distinguishable from human, tuning reward parameters for homogeneous navigation among humans is not trivial \cite{mavrogiannis2017socially}. DRL policies are also a function of the underlying simulator, which often translates to undesirable behavior with the sim-to-real transfer owing to the lack of compatibility with real-world scenarios.

On the other hand, data-driven approaches are common in social trajectory prediction. They aim to directly characterize human interactions observed in the data \cite{rudenko2020human}, eliminating the need for reward shaping and accurate simulations. Recent sequence-to-sequence models, for instance, have achieved promising results in intent prediction \cite{salzmann2020trajectron++, patrikar2022predicting, navarro2022socialpattern}. However, using these models for downstream navigation is difficult as they often suffer from prediction failures \cite{farid2023task} which hurt their generalization capabilities. This potential for unsafe behavior prompts the need for robustifying models deployed in the real world.

\begin{figure}[t]
    \centering
    \includegraphics[width=0.48\textwidth,trim={0cm 0cm -0cm -0cm},clip]{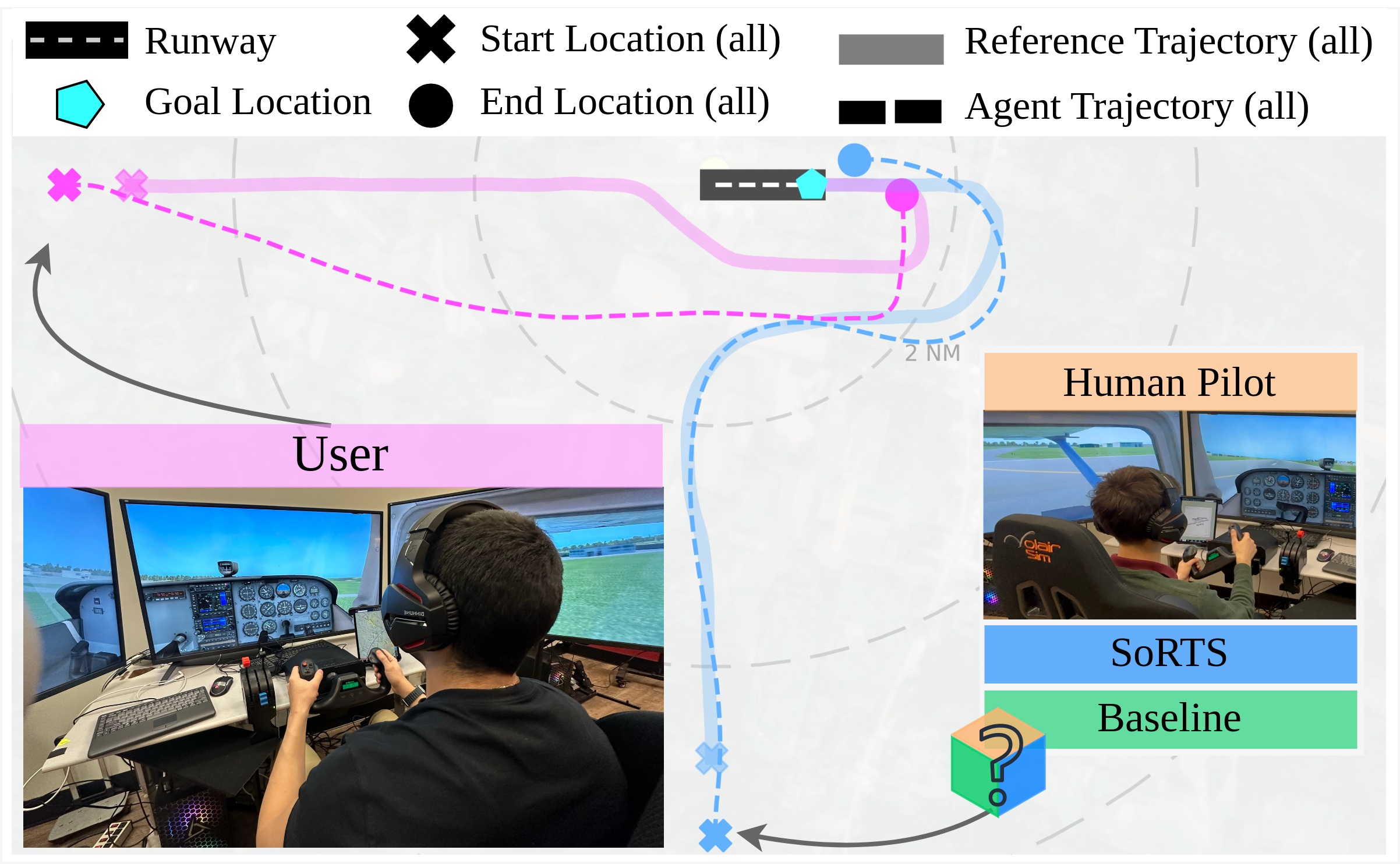}
    \caption{The flight simulator setup used in our user study where, in randomly selected order, each \user~interacted with our \alghuman~pilot, \algsorts~and the \algablation. The figure also shows a resulting interaction between a \user~and \algsorts. }
    \vspace{-0.5cm}
    \label{fig:user_study_setup}
\end{figure}

In social navigation, the actions of one agent influence those of another and vice-versa \cite{sadigh2018planning,hu2022active}. This temporally recursive decision-making intuition has been used for modeling human-like gameplay \cite{silver2018general, jacob2022modeling}. Leveraging this insight, we propose using a recursive \textit{search}-based policy to robustify offline-trained models with downstream \textit{social} navigation objectives. Specifically, we use Monte Carlo Tree Search (MCTS) \cite{kocsis2006bandit} as our \textit{search} policy which provides long-horizon recursive simulations, collision checking and goal conditioning. We combine it with a \textit{socially}-aware intent prediction model to provide short-horizon agent-to-agent context cues and motion naturalness. We use MCTS to fuse these short-horizon cues with long-horizon planning by including global reference paths to guide the tree expansion. We refer to our framework as \textit{Social Robot Tree Search (\sorts)}.


The growing operations of Unmanned Aerial Vehicles are leading to a demand for using airspace concurrently with human pilots \cite{aurambout2019last, grote2022sharing}. We, therefore, select the domain of general aviation (GA) to showcase our approach. GA was recently framed within the paradigm of social navigation \cite{patrikar2022predicting, navarro2022socialpattern}, where pilots are \textit{expected} to follow flying guidelines to coordinate with each other and respect other's personal space to ensure safe operations. This is analogous to following etiquette in human crowds and vehicular settings.

Safety-critical domains, like GA, demand high competence to guarantee seamless and safe operations. This entails developing trustworthy robots which are able to understand and follow navigation norms but also understand long-term dynamic interactions to avoid causing danger or discomfort to others. In this paper, we separate these aspects into two axes, \textit{navigation performance} and \textit{safety}. We center our framework design and evaluations around these two axes. Through a user study conducted on a custom simulator framework, \xplaneros, with 26 experienced pilots, we investigate how pilots interact with our model in a realistic flight setting. We further analyze how they gauge our model along these axes as compared to a competent human pilot. We also complement our evaluations with self-play experiments in scenarios with increasing complexity. 

\textit{Statement of contributions:} 
\begin{enumerate}
    \item We introduce and open-source \sorts, an MCTS-based algorithm for long-horizon navigation that robustifies offline learned socially-aware intent prediction policies for downstream navigation. 
    \item We introduce \xplaneros, a high-fidelity simulation environment for navigation in shared aerial space, and;
    \item Through a user study with 26 FAA-certified pilots and through self-play simulations in more complex scenarios, we show that \sorts~is perceived comparably to a competent human pilot in terms of \textit{navigation performance} and \textit{safety} and significantly outperforms its baseline algorithm. 
\end{enumerate}

%% file: sections/related_works.tex
\section{Related Work} \label{sec:related_works}

\subsection{Social Navigation Algorithms}

Social navigation has a rich body of work focused on pedestrian and autonomous driving domains \cite{mavrogiannis2021core}. In pedestrian settings, classical model-based approaches \cite{berg2011reciprocal, mavrogiannis2022social} have been proposed and remain prominent baselines. Yet, their extension to other domains is non-trivial. Recent DRL methods \cite{matsuzaki2022learning, chen2017decentralized, chen2019crowd} that use handcrafted safety-focused reward functions \cite{tsai2020generative} have produced promising results in these domains.
However, shortcomings in simulator design \cite{biswas2022socnavbench}, and domain-specific reward functions limit real-world performance \cite{tsai2020generative}. Achieving scalability and robustness is challenging, often requiring expensive retraining. 
Similar to our approach, \cite{riviere2021neural} introduces a DRL method that uses MCTS to train and deploy policies. While their method relies on pre-defined reward functions and simulator training, our work extends these methods to use offline expert-based policies, providing domain-specific treatment for social navigation.

Data-driven approaches focus on learning policies from datasets that record interactions between agents \cite{tsai2020generative, mavrogiannis2021core}. As these models do not need explicit reward construction, they can capture the rich, joint social dynamics. However, these methods are challenging to deploy directly owing to noisy and missing demonstrations \cite{bashiri2021distributionally, codevilla2019exploring}. To alleviate this, \cite{sadigh2018planning} used the gradients of a Q-value function for Model Predictive Control, and \cite{hu2022active} proposed a generalization to this method using dual control for belief state propagation. These methods rely on Inverse Reinforcement Learning as an additional step to generate the Q-value functions. Using gradients from sequence models directly in optimizations has also been proposed \cite{schaefer2021leveraging}, but the convergence properties were not examined. Our method is more direct in its use of sequential models and calculating gradients or Q-values is not required. Instead, we transform the model's outputs into action distributions for the downstream planning task. 

\subsection{Social Navigation Evaluation}

Different metrics have been considered for the evaluation of social robot navigation \cite{mavrogiannis2021core, gao2021evaluation, rudenko2020human}. Some of the main axes of analysis for evaluation include behavior naturalness based on a reference trajectory or irregularity of the executed path \cite{mavrogiannis2019effects, sathyamoorthy2020densecavoid}, performance and efficiency \cite{liu2021decentralized, everett2018motion, liang2020realtime}, and notions of physical personal space or discomfort \cite{torta2013design, chen2017socially}. User studies are often conducted to evaluate more subjective aspects such as the perceived discomfort and trust that a robot induces during an interaction \cite{butler2001psychological}. Following prior works, we focus on \textit{navigation performance} to measure our agent's smoothness and ability to follow navigation guidelines, and \textit{safety} to judge its ability to respect others' personal space. We also conduct a user study where we ask experienced pilots to interact with our algorithm in a realistic flight setting and rate the robot's performance, perceived safety and trust. 

%% file: sections/approach.tex
\begin{figure*}[t!]
    \centering
    \includegraphics[width=\textwidth]{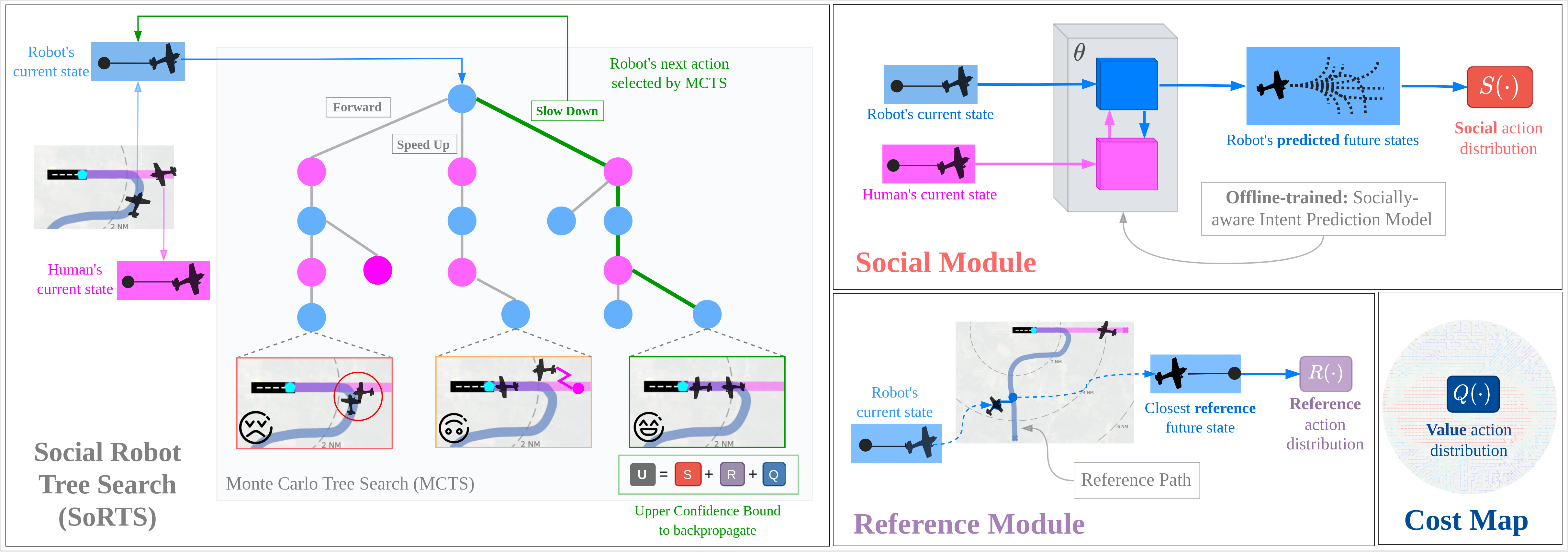}
    \caption{An overview of \sorts, a Monte Carlo Tree Search (MCTS)-based planner for social robot navigation whose tree search is guided by three components; a \modsoc, a \modref~and a \modval. The Social Module leverages an offline-trained intent prediction model to characterize agent-to-agent interactions and predict a set of possible future states for the ego-agent. The Reference Module provides the agent with the closest reference state from a global plan which embodies navigation guidelines the agent must follow. The Cost Map encodes a global visitation map to further guide navigation. MCTS uses all these modules to provide online collision checking and long-horizon simulations by searching for choosing between different decision modalities.}
    \label{fig:model}%
    \vspace{-15pt}%
\end{figure*}

\section{Problem Formulation} \label{sec:problem_formulation}

We formulate social navigation as an approximate POSG, a framework for decentralized finite-horizon planning. For more details about POSGs and their use within social navigation, we refer the reader to \cite{montemerlo2004approximate, riviere2021neural, sadigh2018planning}.Following \cite{navarro2022socialpattern, patrikar2022predicting}, we assume M agents with $\state^{i}_t \in R^{3}$ representing state of agent $i$ at time-step $t$ and $\action^{i}_t \in \mathcal{A}$ is the discrete set of actions or motion primitives that follow $\dot{\state} = f(\state,\action)$. Let $\state^{i}_0$ and $\state^{i}_g$ represent the start and goal states respectively. The system also has access to a set of offline \textit{expert} demonstrations $\mathcal{D}$ and set of global reference trajectories $\tau_R$. We omit superscripts to refer to the joint state for all agents. 

Thus, given the set of start $\state_0$ and goal $\state_g$ states of M agents, the objective is to find a sequence of control inputs $\pi = \{ \action_0, \dots , \action_g \}$ such that the agents follow collision-free trajectories $ \tau = \{ \state, \cdots , \state_g \}$. The generated trajectories need to ensure $||\state^{i}_t - \state^{j}_t|| \geq d ~ \forall i,j \in \{1,\dots,M \} ~ \forall t$ where d is the minimum separation distance to satisfy the safety objective. Furthermore, the trajectories also need to stay close to reference trajectories $\min||\tau-\tau_R||$ to satisfy the navigation objective and follow $\tau \sim \mathcal{D}$ to satisfy the social objective. Without loss of generality, we assume first agent (i=0) to be the robot ego-agent and at each time step execute the optimal action.

\section{Approach} 
\label{ssec:overview}
We designed \sorts~along the axes of \textit{navigation performance} and \textit{safety}. The core insight of SoRTS is to use social modules and global reference paths to bias the MCTS to search for and choose between various decision modalities. Fig. \ref{fig:model} shows an example of two aircraft merging on a single path. Each aircraft can safely execute the merger by choosing from three actions: forward, speeding up, and slowing down. MCTS, in its forward simulations, not only prunes branches that lead to future collisions but also uses the social module to choose between cutting-in-front (socially undesirable) and yielding (socially desirable), thereby producing socially compliant and safe behavior.  

 

\subsection{Modules} \label{ssec:modules}

\sorts~is a search-based planner which uses Monte Carlo Tree Search (MCTS) whose tree expansion is guided by three modules. First, a \textit{Social Module}~ handles the short-horizon dynamics in the scene, characterizing social interactions and cues. Second, a \textit{Reference Module}~ provides the agent with a global navigation guideline, \exempli an airport traffic pattern. Finally, a \textit{Cost Map}~ encodes global value map. MCTS uses these components together and provides collision checking and long-horizon socially-compatible simulations.  Its corresponding pseudo-code is shown in \Cref{algo:sorts_plan} and \ref{algo:sorts_ts}. 


\subsubsection{Social Module} \label{ssec:social} 

We leverage an offline-trained intent prediction algorithm parameterized in $\theta$, $P_\theta(\cdot)$, to account for the short-term agent-to-agent dynamics following the expert trajectories in $\mathcal{D}$.

\begin{equation}
    {p}_\theta(\state_t^i, \action_t^i) \sim P_\theta( {{\action}}_{t}^i \mid {{\state}}_{t-t_{obs}:t}, \state_g^i)
    \label{soc_action}
\end{equation}  

where $p_\theta(\cdot)$ provides a distribution of future actions  ${{\action}_t^i}$ for agent i conditioned on the past trajectories ${{\state}}_{t-t_{obs}:t}$ of all the agents and the goal ${\state_{g}^i}$ where $t_{obs}$ is the observation time horizon.

Here, we use \sprnn~\cite{navarro2022socialpattern}, an algorithm which predicts multi-future trajectories from learned interactions that exploit motion pattern information in the data.

\subsubsection{Reference Path Module} \label{ssec:reference_path}

Given the start and goal state of agent i the algorithm samples a suitable reference trajectory $\tau_r^i \in \tau_{R}$, \cref{algo:reference_path} in \Cref{algo:sorts_plan}. Similar to \cref{ssec:social}, this trajectory is used to compute a \textit{reference} action distribution $p_r(\cdot)$,
\begin{equation}
    {p}_r(\state_t^i, \action_t^i) \sim P_r(\action_t^i | \state_t^i , \tau_r^i )
    \label{ref_action}
\end{equation}  
 proportional to the L2 norm between $\state$ and $\tau_{r}$ at time $t$. In \Cref{algo:sorts_ts}, the reference action is obtained in \cref{algo:reference_action}. $\tau_{R}$ can be drawn from expert distributions $\mathcal{D}$, global path planning algorithms like A-Star, logic specifications \cite{aloor2022follow} or can also be handcrafted. 

\subsubsection{Cost Map} \label{sssec:cost_map}

The algorithm also uses a cost map of the environment representing the \textit{value} function, $v(\state)$. For our case, we use the cost map to represent state visitation frequency to bias the search towards more desirable areas. \Cref{algo:sorts_ts} uses the cost map in \cref{algo:costmap}. This value function can be either learned, \exempli via self-play \cite{riviere2021neural} or pre-computed from a prior distribution and captures the value of the joint state distribution.

\subsection{Social Monte Carlo Tree Search} \label{ssec:planner}

MCTS is a search-based algorithm that expands its tree search toward high rewarding trajectories. In principle, this is done by selecting nodes along the search that maximize upper confidence bound \cite{kocsis2006bandit}, which balances the degree of exploitation and exploration. \sorts, leverages MCTS and uses the components presented in the previous section to guide its trajectory roll-outs. Formally, it follows the UCT shown below,  
\begin{align}
    U(\state, \action) &= 
    Q(\state, \action) + c_1 \cdot S(\state, \action) + c_2 \cdot R(\state, \action) \label{eq:uct} 
\end{align}
where $Q(\state, \action)$ represents the expected value for taking action $\action$ at state $\state$; $S(\state, \action)$ is visitation normalized component according to the socially-aware module; $R(\state, \action)$ is the visitation normalized component according to the reference path; $c_1$ and $c_2$ are hyperparameters. We drop the time $t$ subscript for ease of notation. 
In \cref{algo:update} of \cref{algo:sorts_ts}, these values follow the update:
\begin{align}
    Q(\state, \action) &= \frac{N(\state, \action) \cdot Q(\state, \action) + v(\state) }{N(\state, \action) + 1} \label{eq:Q} \\[0.5em]
    R(\state, \action) &= \frac{N(\state, \action) \cdot R(\state, \action) + p_r(\state, \action) }{N(\state, \action) + 1} \label{eq:PR}\\[0.5em]
    S(\state, \action) &= \frac{\sqrt{N(\state)}}{N(\state, \action)+1} \cdot p_\theta(\state, \action) \label{eq:PS}
\end{align}
where $N(\state)$ is the state visitation count, and $N(\state, \action)$ the visitation given action $\action$. These updates are done iteratively by following the states within a time-budget, or until a new state is found.
At each time-step, a new forward simulation tree is iteratively constructed by alternately expanding the agents' future states in a round-robin fashion. Branches that lead to collision states are pruned.\footnote{Note: In practice, for $M>2$, we only use the ego-agent and the closest agent to the ego agent for tree expansion. While the tree is explicitly constructed only for two agents, $p_\theta$ provides the high-level social context for all the agents. This approximation preserves the real-time nature of the algorithm and is shown to perform well in practice.} At the end of $planHorizon$, the ego agent takes the first action that maximizes the visitation count in \cref{algo:take_action} of \Cref{algo:sorts_plan}. The tree is reset and the process continues till goal is reached. 


\begin{algorithm}
\caption{SoRTS($\state_0, \state_g,  \theta, \tau_R, v$)}
\small
\begin{algorithmic}[1]
\State \colorbox{Orchid!20}{$\tau_r \leftarrow$ GetReferencePaths$(\state_0, \state_g, \tau_R)$} \label{algo:reference_path}
\While{ $\state_t^0 \neq \state^0_g$ or $timeElapsed \le planHorizon$}
\State $N(\cdot) \leftarrow $ SocialMCTS$(\state_t, \state_G, \tau_r, \theta, v, 0)$ \label{algo:call_sorts}
\State $\action^0 \leftarrow \arg\max_{\mathbf{a'} \in \mathcal{A}} N(\state_t^0, \mathbf{a'})$ \label{algo:take_action}
\State $\state_{t+1}^0 \leftarrow f(\state_t^0, \action^0)$ \Comment{Robot's transition model.}
\EndWhile
\end{algorithmic}
\label{algo:sorts_plan}
\end{algorithm}

\begin{algorithm}
\caption{SocialMCTS($\state_t, \state_G, \tau_{r}, \theta, v, p $)}
\small
\begin{algorithmic}[1]
\While{p $\leq$ M}
\If{$\state_t^p \not\in S (\cdot)$ } \Comment{New tree node}
\State \colorbox{Blue!20}{$v_s \leftarrow$ GetValue$(v, \state_t)$} \label{algo:costmap}
\State \colorbox{OrangeRed!20}{$S(\cdot) \leftarrow$ GetSocialActionProbabilities$(\state_t, \theta)$} \label{algo:social_action}
\State \colorbox{Orchid!20}{$p_r(\cdot) \leftarrow$ GetReferenceActionProbabilities$(\tau^p_r, \state_{t}^p)$} \label{algo:reference_action}
\State $N(\state_t^p) \leftarrow 1$
\State \Return $v_s$
\EndIf
\State $\mathcal{A'} \leftarrow$ CollisionCheck$(\mathcal{A},d,\state_t)$
\State $\action^p \leftarrow \arg\max_{\mathbf{a'} \in \mathcal{A'}} U(\state_t^p, \mathbf{a'})$ \Comment{See eq. \ref{eq:uct}}
\State $\state_{t+1}^p \leftarrow f(\state_t^p, \action^p)$
\State p $\leftarrow$ p+1 \Comment{Choose next agent to expand}
\State SocialMCTS$(\state, \state_g, \tau_{r}, \theta$, v, p)
\State Update$(Q, R, S, N)$ \label{algo:update} \Comment{See eq. \ref{eq:Q} to \ref{eq:PS}}
\EndWhile
\State \Return $N(\cdot)$
\end{algorithmic}
\label{algo:sorts_ts}
\end{algorithm}

%% file: sections/experiment_setup.tex
\section{Experimental Setup} \label{sec:setup}

Our experiments focus on assessing \sorts~along our axes of interest, \idest \textit{navigation performance} and \textit{safety}. As such, this section describes the main aspects of our evaluation setup and implementation details for our case study. 

\subsection{User Study} \label{ssec:evaluation_user_study}

We recruited 26 FAA-certified pilots (14 private, 8 commercial, 3 student pilots and 1 airline transport pilot), who on average have 986 flight hours. Using the flight deck setup and simulator shown in \cref{fig:user_study_setup}, each pilot was asked to land an aircraft on a specified runway. Simultaneously, a second pilot was solving the same task, thus, requiring the pilot's coordination to land safely. Here, the second pilot was either a human, \sorts,~or the baseline algorithm\footnote{We use \textit{second pilot} and \textit{algorithm} interchangeably.}.

Our experimental setup follows the design delineated next.
We followed a \textit{within-subject} design in which each user tests against each algorithm. We allowed the user to get familiarized with the simulator and controls prior to the tests.
The pilots are spawned at a 10 km radius from the airport, where their incoming direction is either north (N), south (S), or west (W), defining six possible scenarios, \idest $\{(N, S), (S, N), (N, W), (W, N), (S, W), (W, S)\}$. The algorithm order and scenario were selected randomly at the beginning of the experiment. The scenario, the initial states and final goals remained fixed throughout the three tests. 


After each test, the user completed a 5-scale Likert questionnaire evaluating the second pilot along various factors we deemed relevant for high navigation performance and safety. Since operation in safety-critical domains demands high competence, we also take an interest in understanding which aspects the users prioritize for deeming a pilot as trustworthy and competent. We, thus, ask users to also rate how \textit{trustworthy} and \textit{human-like} they perceived the second pilot, based on their interaction with them and their responses along the axes of navigation and safety. We summarize the components of our user study questionnaire in \cref{tab:user_study_questionaire}. Finally, we also collected the trajectory data from each experiment for further analysis using relevant metrics discussed in \cref{ssec:metrics}. 

\subsection{Self-Play} \label{ssec:self_play_setup}

We complement our user study with self-play simulations to assess \sorts~and our baseline on a wider variety of scenarios. The simulations follow a similar design as the user study, 
but now allow each agent's location to be anywhere around the 10km radius to increase the number of possible scenarios.
We also consider multi-agent scenarios varying from 2 to 4 agents. We randomly generate 100 episodes for each setting, where in each episode an agent is deemed unsuccessful if it breaches a minimum separation distance, or if it reaches a maximum number of allowed steps.

\subsection{Metrics} \label{ssec:metrics}

To quantify the trajectory data collected in our user study and self-play experiments along our axes of analysis, we consider the metrics listed below;
\begin{enumerate}
    \item \textit{Reference Error (RE):} The Euclidean distance between the agent's reference trajectory and its executed path. 
    \item \textit{Loss of Separation (LS):} The duration that two agents break a minimum distance of each other. This metric is relevant for the aviation domain \cite{glozman2021vision}, but is akin to commonly used metrics for social robots in pedestrian settings, \exempli personal space \cite{mavrogiannis2021core, gao2021evaluation}. 
\end{enumerate}

\subsection{Baseline} \label{ssec:ablation}

To showcase the benefits of \sorts, we introduce a baseline which naively chooses the next action by balancing the \textit{reference} and \textit{social} values over the state-action space following the equation below, 
\begin{align*}
    \action = \arg\max_{\action^{'} \in \mathcal{A}} \big[ \lambda \cdot p_r(\state, \action) + (1 - \lambda) \cdot p_\theta(\state, \action)\big]
\end{align*}
where $\lambda \in \R: [0, 1]$ balances the importance we give to $p_\theta(\cdot)$ and $p_r(\cdot)$. This baseline translates to replacing \cref{algo:call_sorts} and \cref{algo:take_action} for the above equation in \cref{algo:sorts_plan}. 

\subsection{Simulator} \label{ssec:simulator}

To evaluate the proposed algorithm and enable future research in the domain of full-scale aerial autonomy, we introduce \xplaneros. The system combines two main components via a ROS bridge, X-Plane-11 and \rosplane~autopilot \cite{ellingson2017rosplane}, enabling the use of high and low-level commands to control a general aviation aircraft in realistic world scenarios. X-Plane-11 is a high-fidelity simulator which provides realistic aircraft models and visuals. \rosplane~is a widely accepted tool which provides reliable autonomous flight control loops. 
Our simulator further adds support for following a select set of motion primitives, as well as visualization utilities that aid in tuning the control loops. The documentation and source code is open-sourced\footnote{Code for \xplaneros: \href{https://github.com/castacks/xplane_ros}{https://github.com/castacks/xplane\_ros}}.


\subsection{Implementation Details} \label{ssec:implementation_details}

The modules in \cref{ssec:modules} leverage \trajair~\cite{patrikar2022predicting}, a dataset consisting of aircraft trajectory data collected in non-towered terminal airspace; the \textit{Social Module} uses \trajair~to train its intent prediction module offline, where we followed the training details in \cite{navarro2022socialpattern}. We also use \trajair~to build a library of FAA-abiding paths used by the \textit{Reference Module}. Finally, to build our \textit{cost map} we discretized \trajair's flight frequency based on aircraft poses and wind direction. 

The action space, $\mathcal{A}$, of our planner is comprised of the set of 252 motion primitives for aerial navigation used in \cite{aloor2022follow}. Empirically, we set $planHorizon$=10s in \cref{algo:sorts_ts} and the exploration parameters in the UCT equation to $c_1=2$ and $c_2=5$, and $\lambda=0.3$ for the baseline planner. 


\begin{table}[!th]
\caption{We asked users to rate the \textit{second pilot} along these aspects of \textit{navigation performance} and \textit{safety}, and to rate it along \textit{trustworthiness} and \textit{humanness}.}
\label{tab:user_study_questionaire}
\resizebox{\columnwidth}{!}{%
\begin{tabular}{@{}ccc@{}}
\toprule
\textbf{Navigation Performance} & \textbf{Safety} & \textbf{Was the second pilot...}\\
\midrule
Follow FAA Guidelines   & Collision Risk   & trustworthy?    \\
Overall Flying Skill    & Comfort          & a human?      \\
Flight Smoothness       & Cooperative      & \\
                        & Abrupt           & \\
                        & Predictable      & \\
 \bottomrule
\end{tabular}}
\vspace{-0.5cm}
\end{table}

%% file: sections/results.tex
\section{Results} \label{sec:results}

\begin{figure*}[t!]
    \centering
    \includegraphics[width=1.0\textwidth]{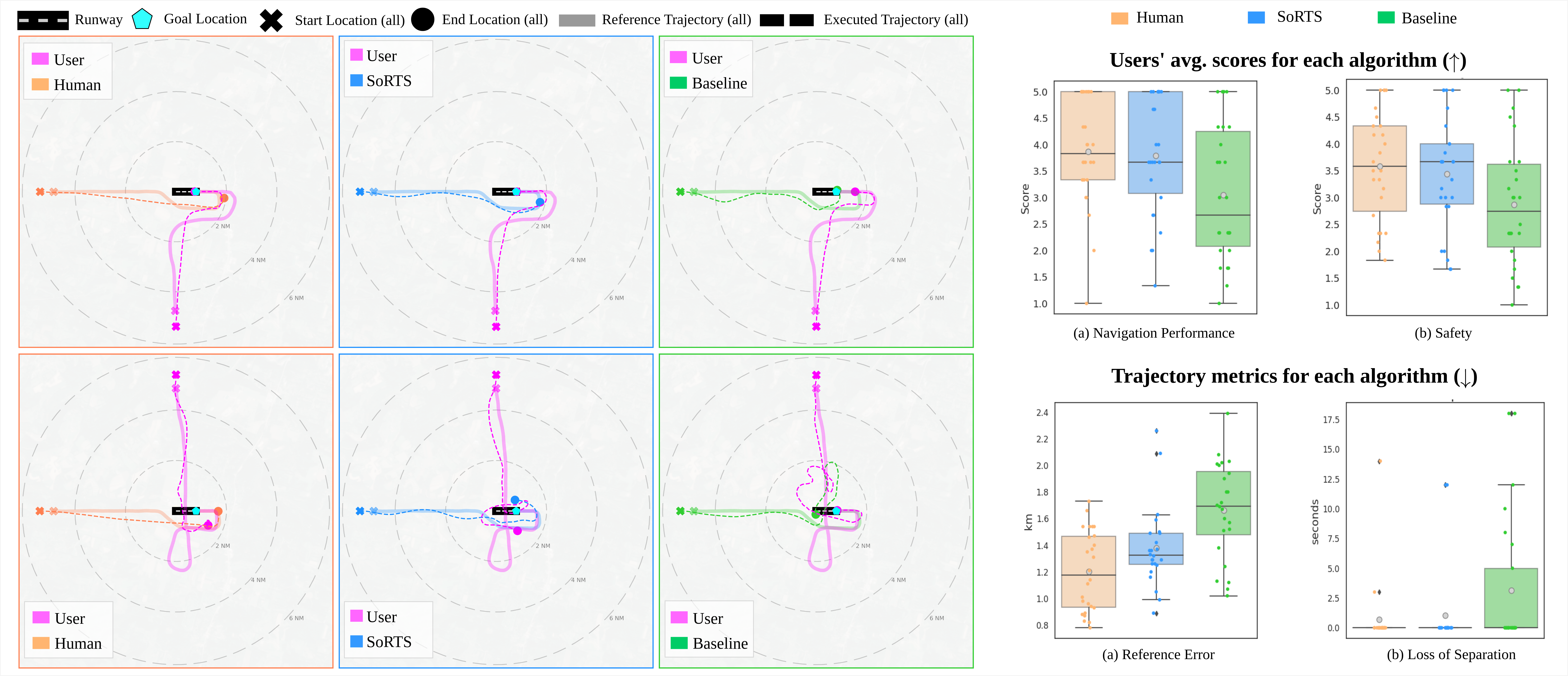}
    \caption{User study results. \textbf{Left:} Examples of two experiments. Each row shows the resulting trajectories of a \user~interacting with our \alghuman~pilot, \algsorts, and the \algablation. The top row shows a user that successfully followed the expected guideline; we can see that the baseline did not fly as smoothly as \sorts, and abruptly cut short close to the goal. The bottom row shows a user that did not follow the expected guideline; our algorithm still managed to navigate properly, whereas the baseline behaved erratically, unsafely crossing over the runway. \textbf{Right:} Box-plots showing the \textit{per-algorithm} distribution of the results from the user study: the top ones show the average scores given by the user for navigation performance (a) and safety (b). The bottom ones show the average reference error (a) and loss-of-separation (b) metrics obtained from the trajectory data.}
    \label{fig:user_study}
    \vspace{-15pt}%
\end{figure*}

We now present our results and insights. For our analysis below, a \textit{competent algorithm} refers to one which performs highly along the axes of navigation performance and safety. We first examine the relevance of the factors in \cref{tab:user_study_questionaire} in characterizing the \textit{trustworthiness} of a competent algorithm. Leveraging the results from our \textit{trustworthiness} analysis, we then perform a comparative analysis between \sorts, the baseline and our human pilot.  Finally, we briefly explore how pilots perceive human competence along the factors in our questionnaire.

\subsection{On competence and trustworthiness} \label{ssec:trust}

As a means to determine how to better assess the competence of a robot, \exempli which metrics to prioritize, we compare the factors in \cref{tab:user_study_questionaire} with the user's perceived trust. We do so via a Pearson's correlation analysis with repeated measures, whose results are summarized in \cref{tab:us_rm_corr}. 

We find strong correlations between trust and the components within navigation performance, with flight smoothness being the strongest one. 
Within safety, we see that \textit{comfort}, followed by \textit{predictability} showed the strongest correlations. Although one would expect for \textit{abruptness} and \textit{collision risk} to be as crucial for rating trust. We find this result is, in part, due to our experiments not being explicitly designed to exhibit adversarial behavior. We also believe these results showcase the strength of our model in unprecedented scenarios. As said earlier, because data-driven models can heavily misbehave when exposed to out-of-distribution states \cite{farid2023task}, \sorts~is a means to augment these models online. Since, the user study serves as a mechanism for testing our model in settings that weren't observed in the training data, the lower correlation between the aforesaid components may indicate robustness toward unsafe behavior.

\begin{table}[!ht]
\centering
\caption{Using repeated measures, we show the Pearson correlations (R, p-value=0.05) for each factor listed in \cref{tab:user_study_questionaire} \textit{vs.} perceived trustworthiness and humanness.}
\label{tab:us_rm_corr}
\resizebox{\columnwidth}{!}{%
\begin{tabular}{@{}clcccc@{}}
\toprule
 \multirow{2}{*}{Axis} & \multirow{2}{*}{Factor} & \multicolumn{2}{c}{Trustworthy?} & \multicolumn{2}{c}{Human?} \\
 \cmidrule(l){3-6} 
 & & R & p-value & R & p-value \\ \midrule
\multirow{3}{*}{Nav. Performance}   & Flight Smoothness & 0.81 & 4.80e-19 & 0.08 & 0.47 \\ 
 & Follow FAA Guidelines & 0.76 & 2.99e-15 & 0.10 & 0.37 \\
 & Overall Flying Skill & 0.71 & 4.56e-13 & 0.15 & 0.20 \\
\midrule
\multirow{5}{*}{Safety}   & Comfortable & 0.92 & 3.55e-31 & 0.17 & 0.14 \\
  & Predictable Behavior & 0.77  & 5.19e-16 & 0.21 & 0.07 \\
  & Cooperative & 0.65 & 1.61e-10 & 0.06 & 0.62 \\
& Collision Risk & -0.56 & 1.10e-07 & -0.11 & 0.36 \\
  & Abrupt      & -0.56 & 1.13e-07 & -0.09 & 0.42 \\
\bottomrule
\end{tabular}
}
\end{table}

\subsection{On each algorithm's competence}

The previous section studied how each factor in \cref{tab:user_study_questionaire} tied to the perception of a trustworthy pilot. This section will now focus on providing a pairwise comparison between the \textit{algorithms} for both the user study and self-play, leveraging the results obtained in the previous analysis.

\subsubsection{\textbf{User study}}

Leveraging the trustworthiness results in \cref{tab:us_rm_corr}, we compute each user's average score over the questions with highest correlations along each axis, \idest for navigation performance, we compute the mean score between \textit{following FAA} and \textit{flight smoothness}, and for safety we use \textit{predictability} and \textit{comfort}. The scores are shown in \cref{fig:user_study}, and the pairwise statistical differences using ANOVA with repeated measures in \cref{tab:us_statistical_analysis}. Here, we did not find statistical evidence that the scores for the human pilot and \sorts~were different, suggesting that the users rated their competence similarly. Our results further show evidence that our baseline was generally rated lower on both axes of competence, while also exhibits more variance. 

We then examine if the RE and LS metrics in \cref{ssec:metrics}—which here tie to navigation performance and safety, respectively—reflect the users' assessments for each algorithm. Thus, we compute them on the collected trajectories and show their respective mean scores and statistical comparisons in \cref{fig:user_study} and \cref{tab:us_statistical_analysis}.
As before, we show the mean scores for each algorithm in \cref{fig:user_study} and the statistical results in \cref{tab:us_statistical_analysis}. 
For the RE metric, we find significantly different results between the algorithms; wherein \sorts~yields higher error compared the human pilot, but markedly lower than the baseline and less variance than the other two. Similarly, the LS metric computed at 0.5km, shows that generally neither the human pilot nor \sorts~breach this distance. In contrast, the baseline more frequently invades others' personal space, creating more situations for potential collisions. 

\begin{table}[!ht]
\centering
\caption{Statistical significance between algorithmic pairs for results in \cref{fig:user_study} with $t^\ast \geq 2.060$ and $p\leq0.05$.}
\label{tab:us_statistical_analysis}
\resizebox{\columnwidth}{!}{%
\begin{tabular}{@{}cc@{\hspace{0.2cm}}cc@{\hspace{0.2cm}}cc@{\hspace{0.2cm}}cc@{\hspace{0.2cm}}c@{}}
\toprule
\multirow{2}{*}{\textbf{Algorithmic}} & \multicolumn{2}{c}{\textbf{NP}} & \multicolumn{2}{c}{\textbf{Safety}} & \multicolumn{2}{c}{\textbf{RE}} & \multicolumn{2}{c}{\textbf{LS}} \\ \cmidrule(l){2-9} 
 \textbf{Pair} & t-val & p-val & t-val & p-val & t-val & p-val & t-val & p-val \\ \midrule
Baseline-Human & 3.121 &  0.009 & 3.062 & 0.016 & 5.782 & 0.000 & 1.321 & {\color{OrangeRed} 0.199} \rule{0pt}{2.6ex}\\
Baseline-\sorts & 3.018 &  0.009 & 2.626 &  0.022 & 2.105 & 0.011 & 0.397 & {\color{OrangeRed} 0.694} \rule{0pt}{2.6ex}\\
Human-\sorts & {\color{OrangeRed} 0.415} & {\color{OrangeRed} 0.682} & {\color{OrangeRed} 0.322} & {\color{OrangeRed} 0.322} & 2.944 & 0.022 & 1.211 & {\color{OrangeRed} 0.237}  \rule{0pt}{2.6ex} \\ 
\bottomrule
\end{tabular}
}
\begin{tablenotes}
\scriptsize
\item \textbf{NP}: Navigation Performance, \textbf{LS}: Loss of Separation, \textbf{RE}: reference error. 
\end{tablenotes}
\vspace{-0.2cm}
\end{table}

\Cref{fig:user_study} shows examples of trajectories from our experiments. Each row represents one user \textit{vs.} the three \textit{algorithms}. We show a reference trajectory along with the actual executed trajectory\footnote{Human pilots do not see the reference trajectory, but they see the map in the simulator, and are instructed to follow the appropriate traffic patterns.}. The top row shows an example where the user performed successfully. In this experiment, we see that the ablation unexpectedly cut short while approaching the runway instead of following the traffic pattern, while \sorts~did so smoothly and safely. The bottom row shows a user that did not follow the standard traffic pattern. Despite this, \sorts~manages to successfully complete the task, while the baseline behaves erratically, not following the traffic pattern and unsafely traversing the runway twice.  

As per the user study, we conclude that \sorts~performs comparable to a competent human pilot and significantly better than the ablation, both as perceived by the users and through our metrics. Our results further highlight the benefits of using the long-horizon socially-aware simulations via MCTS, as opposed to purely using a data-driven model on the wild with a simple weighting over the objectives of interest, leading to an erratic and unsafe behavior. 

\subsubsection{\textbf{Self-play}}


\begin{table}[!th]
\centering
\caption{Summary of the task performance of the baseline and \sorts' agents in self-play.}
\label{tab:self-play_results}
\begin{tabular}{@{}cccccc@{}}
\toprule
\multirow{2}{*}{\textbf{\# Agents}} & \multirow{2}{*}{\textbf{Algorithm}} & \multirow{2}{*}{\textbf{Success (\%)}} & \multicolumn{2}{c}{\textbf{Failure (\%)}} & \multirow{2}{*}{\textbf{RE}} \\
 &  &  & \textbf{LS} & \textbf{Timeout} & \\ 
\midrule
\multirow{2}{*}{2} & Baseline & 69.0 & 30.0 & 1.0 & 1.59 \rule{0pt}{2.6ex}\\
& \sorts & 96.5 & 2.0 & 1.5 & 2.01 \rule{0pt}{2.6ex}\\
\midrule
\multirow{2}{*}{3} & Baseline & 48.3 & 49.3 & 2.4 & 1.55 \rule{0pt}{2.6ex}\\
& \sorts & 89.7 & 8.3 & 2.0 & 2.01 \rule{0pt}{2.6ex}\\
\midrule
\multirow{2}{*}{4} & Baseline & 43.5 & 54.0 & 2.5 & 1.56 \rule{0pt}{2.6ex}\\
& \sorts & 71.0 & 21.0 & 8.0 & 1.96 \rule{0pt}{2.6ex} \\
\bottomrule
\end{tabular}
\begin{tablenotes}
\scriptsize
\item \textbf{LS}: Loss of Separation at 0.5km, \textbf{RE}: reference error for successful agents. 
\end{tablenotes}
\vspace{-0.3cm}
\end{table}

We now analyze the performance of our algorithm in more complex scenarios. The results summarized in  \Cref{tab:self-play_results} show the percentage of agents that were able to land on the runway, \idest successful agents, the agents that were unsuccessful, due to either a loss of separation, or because they exceeded the maximum allowed time. Finally, we show the average reference error for successful agents.  

Analyzing the table, we observe a decrease in task success as the number of agents increases. Nonetheless, we can see that \sorts~performs significantly better than the ablation algorithm improving the task success by 28\%, 46\% and 38\% for the 2, 3 and 4 agent scenarios, respectively. We also find that, the average reference error for \sorts~is higher than the ablation. Though this error was computed on successful agents only, we hypothesize that it being higher is due to conflict resolution to avoid collisions with other agents. 


\subsection{On competence and human performance}

Often arising debate, is the question of whether an algorithm can pass as a human. However, in domains such as aviation, where high competence is central, one would expect for humans to be perceived as such. If so was the case, one would strive for an algorithm to exhibit a performance similar to that of a human. We study the aforesaid by asking users to gauge whether the second pilot was a human based on their assessments along \cref{tab:user_study_questionaire}. 

Surprisingly, we find a marked disagreement within the users' responses to this question, with almost a 50-50\% split between the responses for the \textit{Human Pilot}~(14: No, 12: Yes) and the \textit{Baseline}~(14: No, 12: Yes), whereas for {\it\sorts}~(8: No, 18: Yes) more users perceived its performance as human-like. Further, \cref{tab:us_rm_corr} shows that the user's assessments for navigational performance and safety correlate weakly with the \textit{humanness} prediction. To explain this contrasting result, we isolated the responses given to \sorts~from the other two, and found that \textit{predictability} had the highest correlation value among all factors, with R=0.53 and p=0.01. In contrast, the values for the Human pilot and the Baseline were R=0.22, p=0.03 and R=-0.06, p=0.06, respectively. We believe this result may suggest that because users generally perceived the behavior of \sorts~as more predictable, they rated its performance as human-like more frequently. 

We believe this contrasting result does not affect our analyses from the previous to sections, as we find that the user's responses for \textit{trustworthiness} also correlate weakly with their assessment of \textit{humanness}, with R=0.26, p=0.02. As such, we conclude that users value more their perception of trust than the nature of type of agent they interact with. 

%% file: sections/conclusion.tex
\section{Conclusion} \label{sec:conclusions}


We present \sorts, an algorithm for long-horizon social robot navigation. \sorts~is a MCTS-based planner which expands its search tree guided by an offline-trained intent prediction model and a global path which embodies navigation guidelines. We introduce \xplaneros, a high fidelity simulator for research in full-scale aerial autonomy. We use it to conduct a user study with experienced pilots to study our algorithm's performance in realistic flight settings. We find that users perceive \sorts~comparable to a competent human pilot and significantly better than our baseline. In self-play, we show that \sorts~outperforms the baseline by 28-46\% as the complexity of scenarios increases. To the best of our knowledge, this is the first work in social navigation for general aviation and attempts to bring unique problems in general aviation within the purview of the larger robotics community. 


We identify various avenues for future work. First, we assumed task homogeneity, \idest agents landing on the same runway, whereas in a real scenario pilots with different objectives may require to interact. Thus, future work includes studying interactions with agents with heterogeneous tasks. We also assumed perfect intent and state estimation. 
Accordingly, robustifying prediction models with uncertainty- and adversarial-awareness is another promising direction. 
Finally, improvements on the scalability of the model, as well as exploring other domains are also potential avenues. 
